\documentclass{article}

\usepackage{microtype}
\usepackage{graphicx}
\usepackage{subfigure}
\usepackage{booktabs} 

\usepackage{makecell}

\usepackage{hyperref}

\usepackage{tikz}
\usepackage[edges]{forest}
\usepackage{graphicx}

\usepackage{xcolor}        
\usepackage{colortbl}

\usepackage{tcolorbox} 
\usepackage{enumitem}   

\definecolor{hidden-red}{RGB}{205, 44, 36}
\definecolor{hidden-blue}{RGB}{194,232,247}
\definecolor{hidden-orange}{RGB}{243,202,120}
\definecolor{hidden-green}{RGB}{34,139,34}
\definecolor{hidden-pink}{RGB}{255,245,247}
\definecolor{hidden-black}{RGB}{20,68,106}
\definecolor{purple}{RGB}{144,153,196}
\definecolor{yellow}{RGB}{255,228,123}
\definecolor{hidden-yellow}{RGB}{255,248,203}
\definecolor{tkcolor}{RGB}{224,223,255}
\definecolor{darkblue}{rgb}{0, 0.40, 0.75}
\newcommand{\eg}{\textit{e.g.,}}
\hypersetup{colorlinks=true, citecolor=darkblue, linkcolor=darkblue, urlcolor=darkblue}
\tcbset{
  aibox/.style={
    width=\linewidth,
    top=8pt,
    bottom=4pt,
    colback=blue!6!white,
    colframe=black,
    colbacktitle=black,
    enhanced,
    center,
    attach boxed title to top left={yshift=-0.1in,xshift=0.15in},
    boxed title style={boxrule=0pt,colframe=white,},
  }
}

\newtcolorbox{AIbox}[2][]{aibox,title=#2,#1}

\usepackage[accepted]{icml2025}

\usepackage{amsmath}
\usepackage{amssymb}
\usepackage{mathtools}
\usepackage{amsthm}

\usepackage[capitalize,noabbrev]{cleveref}

\theoremstyle{plain}

\theoremstyle{definition}

\theoremstyle{remark}

\usepackage[textsize=tiny]{todonotes}

\icmltitlerunning{Towards Geometry Problem Solving in the Large Model Era: A Survey}

\begin{document}

\twocolumn[
\icmltitle{Towards Geometry Problem Solving in the Large Model Era: A Survey}

\icmlsetsymbol{equal}{*}

\begin{icmlauthorlist}
\icmlauthor{Yurui Zhao}{yyy}
\icmlauthor{Xiang Wang}{yyy}
\icmlauthor{Jiahong Liu}{zzz}
\icmlauthor{Irwin King}{zzz}
\icmlauthor{Zhitao Huang}{yyy}
\end{icmlauthorlist}

\icmlaffiliation{yyy}{College of Electronic
Science and Technology, National University of Defense Technology,
Changsha, China}
\icmlaffiliation{zzz}{Department of Computer Science and Engineering, Chinese University of Hong Kong, Hong
Kong, China}

\icmlcorrespondingauthor{Xiang Wang}{xwang@nudt.edu.cn}
\icmlkeywords{Math Reasoning, Geometry Problem Solving, LLM}
\vskip 0.3in
]



\printAffiliationsAndNotice{}  

\begin{abstract}
Geometry problem solving (GPS) represents a critical frontier in artificial intelligence, with profound applications in education, computer-aided design, and computational graphics.
Despite its significance, automating GPS remains challenging due to the dual demands of spatial understanding and rigorous logical reasoning. 
Recent advances in large models have enabled notable breakthroughs, particularly for SAT-level problems, yet the field remains fragmented across methodologies, benchmarks, and evaluation frameworks.
This survey systematically synthesizes GPS advancements through three core dimensions: (1) benchmark construction, (2) textual and diagrammatic parsing, and (3) reasoning paradigms. We further propose a unified analytical paradigm, assess current limitations, and identify emerging opportunities to guide future research toward human-level geometric reasoning, including automated benchmark generation and interpretable neuro-symbolic integration.
\end{abstract}

\section{Introduction}
\label{section:intro}

Geometry problem solving (GPS) has long posed a persistent challenge in mathematical reasoning and artificial intelligence research \cite{bobrow1964natural, chou1996automated}. Successfully automating GPS demands three core capabilities: (1) parsing geometric information from diagrams and textual descriptions to extract spatial relationships \cite{seo2015solving, zhang2022plane}, 
(2) constructing logical reasoning chains to deduce stepwise solutions \cite{itzhaky2013solving, chen2021geoqa, chen2022unigeo}, and 
(3) performing numerical calculations to derive precise answers \cite{lu2021inter, chen2022unigeo}. It is worth noting that merging logical reasoning and numerical calculation in GPS is common because geometric concepts inherently link to numbers, enabling iterative verification, unified symbolic-numerical representation, and supporting automated geometric modeling.

\begin{figure}[!h]
    \centering
    \includegraphics[width=\linewidth]{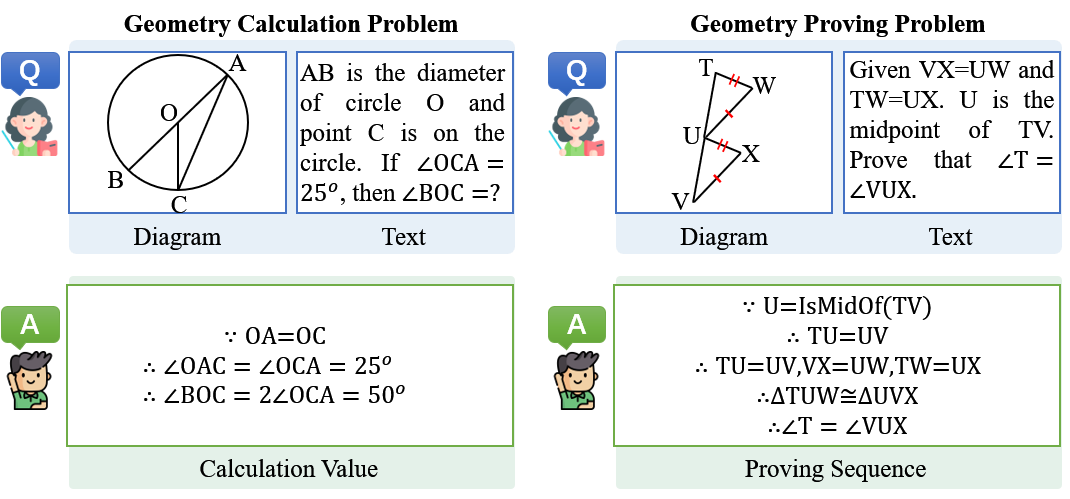}
    \caption{Schematic plot of GPS, which contains geometry calculation problem and geometry proving problem.}
    \label{fig:task}
\end{figure}

Early research in GPS primarily focused on symbolic solvers \cite{seo2015solving, sachan2017learning}, which relied on computationally intensive search algorithms and basic deductive reasoning. 
Though these systems achieved progress in constrained settings, their dependence on rigid rule-based frameworks limited applicability to elementary problems (Grade 6–10 level). 
Recent continuous breakthroughs in artificial intelligence technologies, especially with the coming of the large model era \cite{liu2023visual, bai2023qwen, achiam2023gpt, grattafiori2024llama, liu2025survey}, propel GPS into a new wave of advancement.
Advances in GPS methodologies are now bridging these historical limitations, positioning the technology as a foundational tool for emerging interdisciplinary applications, including automated theorem proving \cite{loveland2016automated}, computer-aided design (CAD) \cite{bi2020computer}, and AI-driven educational systems \cite{lin2023artificial}.

\begin{figure*}[!h]
    \centering
    \includegraphics[width=0.85\linewidth]{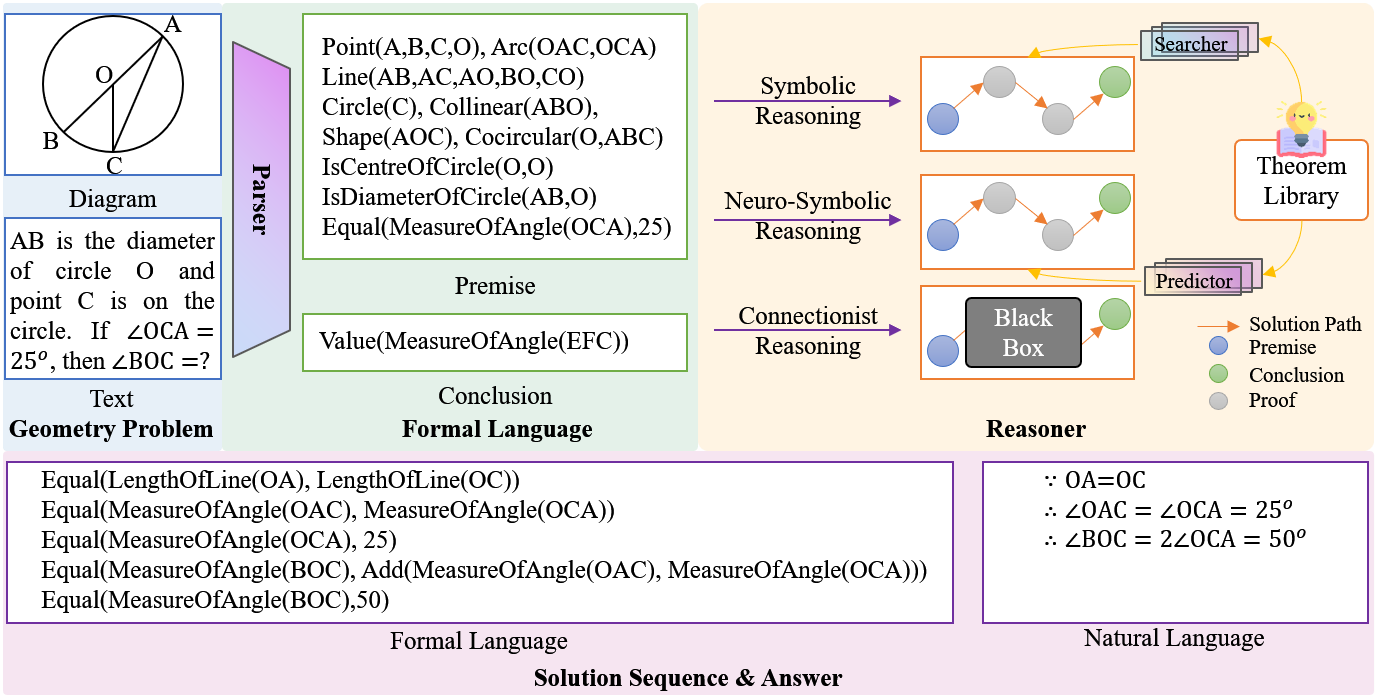}
    \caption{Paradigm of geometry problem solving.}
    \label{fig:framework}
\end{figure*}

However, the inherent complexity of GPS introduces critical challenges that impede system performance across three key areas: benchmark construction, information parsing, and logical reasoning. 
First, creating high-quality, fine-grained, and large-scale benchmarks demands extensive expert annotation \cite{pan2025enhancing,deng2024r}, limiting dataset scalability.
Second, parsing geometric primitives requires accurately interpreting complex layouts and relationships between visual elements \cite{zhang2022plane,zhang2025diagram}.
Third, logical reasoning frameworks must balance interpretability with reasoning efficiency \cite{lu2021inter,ning2025gns}, especially in zero-shot cases.

The current GPS research landscape, marked by diverse methodologies, fragmented technical approaches, and heterogeneous evaluation benchmarks, has driven innovation but also created a disconnected knowledge base. To address this, we systematically collect, analyze, and synthesize the existing literatures, offering the GPS community a unified perspective and actionable roadmap.
The main contributions of this survey are summarized as follows:
\textbf{(1) Comprehensive Survey}: To the best of our knowledge, this represents the first systematic survey dedicated specifically to GPS, providing the research community with essential foundational knowledge and analytical insights.
\textbf{(2) Structured Taxonomy}: 
We propose a three-layer taxonomy for current literature addressing core technical challenges, i.e., benchmark construction, parsing, and reasoning.
\textbf{(3) Unified Reasoning Paradigm}: We develop a systematic paradigm integrating symbolic reasoning, connectionist reasoning, and neuro-symbolic reasoning into a unified paradigm for GPS.
\textbf{(4) Summary of Benchmarks:}  We analyze benchmark design methodologies and highlight automated construction as a pivotal direction for scalable, high-quality dataset synthesis.

\section{Problem Description and Paradigm}
\label{section:GPS}
\subsection{Problem Statement}

\tikzstyle{my-box}=[
rectangle,
draw=hidden-black,
rounded corners,
text opacity=1,
minimum height=1.5em,
minimum width=5em,
inner sep=2pt,
align=center,
fill opacity=.5,
]
\tikzstyle{leaf}=[
my-box, 
minimum height=1.5em,
fill=hidden-blue!57, 
text=black,
align=left,
font=\normalsize,
inner xsep=5pt,
inner ysep=4pt,
align=left,
text width=45em,
]
\tikzstyle{leaf2}=[
my-box, 
minimum height=1.5em,
fill=green!15, 
text=black,
align=left,
font=\normalsize,
inner xsep=5pt,
inner ysep=4pt,
]
\tikzstyle{leaf3}=[
my-box, 
minimum height=1.5em,
fill=yellow!32, 
text=black,
align=left,
font=\normalsize,
inner xsep=5pt,
inner ysep=4pt,
]
\tikzstyle{leaf4}=[
my-box, 
minimum height=1.5em,
fill=green!20, 
text=black,
align=left,
font=\normalsize,
inner xsep=5pt,
inner ysep=4pt,
]

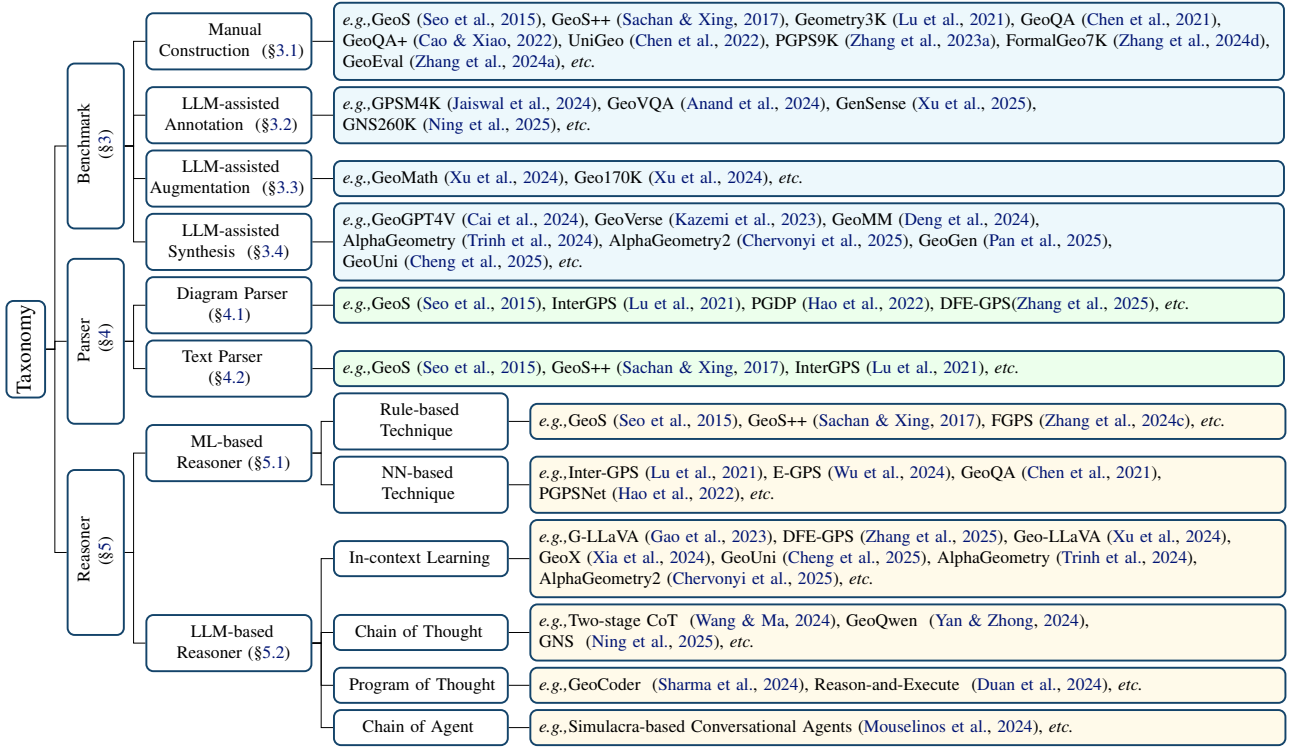
\begin{figure*}[!t]
\vspace{-2mm}
\centering
\resizebox{\textwidth}{!}{
	\begin{forest}
		for tree={
                forked edges,
			grow=east,
			reversed=true,
			anchor=base west,
			parent anchor=east,
			child anchor=west,
			base=left,
			font=\large,
			rectangle,
			draw=hidden-black,
			rounded corners,
			align=left,
			minimum width=4em,
			s sep=3pt,
			inner xsep=2pt,
			inner ysep=4pt,
			line width=1.1pt,
			ver/.style={rotate=90, child anchor=north, parent anchor=south, anchor=center},
		},
		where level=1{text width=9em,font=\normalsize,}{},
        where level=2{text width=9em,font=\normalsize,}{},
        where level=3{text width=9.5em,font=\normalsize,}{},
        where level=4{text width=52em,font=\normalsize,}{},
[Taxonomy, ver
	[\ \ \ \ \ \ \ Benchmark \\  \hspace{3.1em} ~(\S\ref{section:benchmark}),ver
        [\hspace{3em} Manual \\ \hspace{0.2em} Construction ~(\S\ref{section:data_manu})
                [\eg GeoS~\citep{seo2015solving}{,} GeoS++~\citep{sachan2017learning}{,} Geometry3K~\citep{lu2021inter}{,}  GeoQA~\citep{chen2021geoqa}{,} \\
                GeoQA+~\citep{cao2022augmented}{,} UniGeo~\citep{chen2022unigeo}{,} 
                PGPS9K~\citep{zhang2023multi}{,}
                FormalGeo7K~\citep{zhang2024formal}{,}\\
                GeoEval~\citep{zhang2024geoeval}{,}
                \textit{etc.}
				, leaf, text width=53.1em]]
			[\hspace{1.5em} LLM-assisted \\ \hspace{0.5em} Annotation ~(\S\ref{section:data_llm_annotation})
				[\eg GPSM4K~\citep{jaiswal2024advancing}{,} GeoVQA~\citep{anand2024geovqa}{,} 
                GenSense~\cite{xu2025geosense}{,}\\
                GNS260K~\cite{ning2025gns}{,}
                \textit{etc.}, leaf, text width=53.1em]
			]	
			[\hspace{1.5em} LLM-assisted \\ Augmentation ~(\S\ref{section:data_llm_aug}) 
				[\eg GeoMath~\cite{xu2024geo}{,}
                Geo170K~\citep{xu2024geo}{,} \textit{etc.}
				, leaf, text width=53.1em]
			]
		[\hspace{1.5em} LLM-assisted \\ \hspace{0.7em} Synthesis ~(\S\ref{section:data_llm_syn}) 
			[\eg GeoGPT4V~\citep{cai2024geogpt4v}{,}
                GeoVerse~\citep{kazemi2023geomverse}{,}
                GeoMM~\citep{deng2024r}{,} \\
                AlphaGeometry~\cite{trinh2024solving}{,}
                AlphaGeometry2~\cite{chervonyi2025gold}{,}
                GeoGen~\cite{pan2025enhancing}{,}\\
                GeoUni~\cite{cheng2025geouni}{,}
                 \textit{etc.}
			, leaf, text width=53.1em]
		]  
	]
	[\hspace{3em} Parser \\ \hspace{3.2em} ~(\S\ref{section:parser}), ver
		[\hspace{1.2em} Diagram Parser \\ \hspace{3em}~(\S\ref{section:paser_diagram})
        	[\eg GeoS~\citep{seo2015solving}{,} 
                InterGPS~\citep{lu2021inter}{,} 
                PGDP~\citep{hao2022pgdp5k}{,} 
                DFE-GPS\citep{zhang2025diagram}{,} 
                \textit{etc.}
		      , leaf2, text width=53.1em]
		]        
		[\hspace{1.5em} Text Parser \\ \hspace{3em}~(\S\ref{section:paser_text})
        	[\eg GeoS~\citep{seo2015solving}{,} 
                GeoS++~\citep{sachan2017learning}{,} 
                InterGPS~\citep{lu2021inter}{,} 
                \textit{etc.}
				, leaf2, text width=53.1em]
		]
	]
	[\hspace{2.5em} Reasoner \\ \hspace{3.1em} ~(\S\ref{section:reason}), ver
    [\hspace{2em}  ML-based \\ \hspace{1.2em}  Reasoner (\S\ref{section:ml})
		[\hspace{2.1em} Rule-based \\ \hspace{2.1em} Technique
			[\eg 
            GeoS~\cite{seo2015solving}{,}
            GeoS++~\cite{sachan2017learning}{,}
            FGPS~\cite{zhang2024fgeo}{,}
            \textit{etc.}, leaf3, text width=42em]
		]
		[\hspace{2.2em} NN-based \\ \hspace{2.2em} Technique
			[\eg Inter-GPS~\cite{lu2021inter}{,}
            E-GPS~\cite{wu2024gps}{,}
            GeoQA~\cite{chen2021geoqa}{,}\\
            PGPSNet~\cite{hao2022pgdp5k}{,}
            \textit{etc.}, leaf3, text width=42em]
		]
	]
	[\hspace{2em} LLM-based \\ \hspace{1.2em} Reasoner  (\S\ref{section:llm})
		[\ \  In-context Learning
		[\eg G-LLaVA~\citep{gao2023g}{,}
        DFE-GPS~\citep{zhang2025diagram}{,}
        Geo-LLaVA~\citep{xu2024geo}{,}\\
        GeoX~\citep{xia2024geox}{,}
        GeoUni~\citep{cheng2025geouni}{,}
        AlphaGeometry~\citep{trinh2024solving}{,}\\
        AlphaGeometry2~\citep{chervonyi2025gold}{,}
        \textit{etc.}, leaf3, text width=42em]
		]
		[\ \ \ Chain of Thought
			[\eg Two-stage CoT ~\citep{wang2024mutli}{,}
            GeoQwen ~\citep{shengyuan2024geo}{,} \\
            GNS ~\citep{ning2025gns}{,}
            \textit{etc.} , leaf3, text width=42em]
		]
		[\ \ Program of Thought
			[\eg  GeoCoder ~\citep{sharma2024geocoder}{,}
            Reason-and-Execute ~\citep{duan2024reason}{,} \textit{etc.} 
		, leaf3, text width=42em]
		]  
		[\hspace{1.1em} Chain of Agent
		[\eg Simulacra-based Conversational Agents \citep{mouselinos2024beyond}{,}  
            \textit{etc.}, leaf3, text width=42em]
		]    		
        ]
    ]    
]
\end{forest}
}
\caption{Taxonomy of Geometry Problem Solving.\label{fig:taxonomy}}
\end{figure*}

GPS encompasses the computational resolution of geometric problems, which fundamentally divide into two categories (as illustrated in Fig.~\ref{fig:task}): calculation problems determine quantitative measurements like lengths, areas, and angles \cite{seo2015solving,chen2021geoqa}, while proof problems establish geometric truths through deductive arguments \cite{trinh2024solving,chervonyi2025gold}.
The left panel of Fig.~\ref{fig:task} exemplifies a calculation problem: applying the central angle theorem (where an inscribed angle subtending an arc is half its central angle), we derive $\angle BOC = 2\angle OCA$ to compute $\angle BOC = 50^\circ$. The right panel demonstrates a proof problem: establishing triangle congruence via the Side-Side-Side (SSS) criterion to deduce equality of corresponding angles.

Though distinct in objectives, these problem types exhibit deep interconnections: calculation techniques often rely on proof-derived theorems, while computational results provide intuition guiding formal proofs \cite{chen2022unigeo,trinh2024solving}. Given this inherent interdependence, modern computational frameworks increasingly adopt integrated approaches that unify both problem types within cohesive GPS systems.
Formally, we represent the GPS task as
\begin{equation}
    ({\cal S},{\cal A}) = GPS({\cal X} \oplus {\cal Y},{\cal Q},{\cal T}),
    \label{eq:abstract}
\end{equation}
where $\cal X$ and $\cal Y$ denote problem statements in diagrams and text, respectively. 
$\cal T$ represents the external knowledge library. 
$\cal S$ and $\cal A$ are respectively the solution sequence and the answer for the question $\cal Q$.

\subsection{GPS Paradigm\label{section:paradigm}}
We propose a systematic paradigm for GPS that encompasses two fundamental stages: parsing and reasoning, as illustrated in Fig.~\ref{fig:framework}. 
This paradigm can be formally represented as:
\begin{equation}
({\cal S},{\cal A}) = g(f({\cal X} \oplus {\cal Y}, {\cal Q}), {\cal T}),
\end{equation}
where $f$ and $g$ denote the parsing and reasoning function respectively.  

\textbf{Parsing Stage}: Given a geometry problem $({\cal X}  \oplus {\cal Y}, {\cal Q})$, the parsing component is to design a function $f$ that transforms the multimodal input into formal languages: $(\tilde{\cal X} \oplus \tilde{\cal Y}, \tilde{\cal Q}) = f({\cal X} \oplus {\cal Y}, {\cal Q})$, where $\tilde{\cal X}$, $\tilde{\cal Y}$, and $\tilde{\cal Q}$ are in the formal language space $\Omega$.

\textbf{Reasoning Stage}: The reasoning component $g$ constructs a valid logical chain $\cal S$ and calculates accurate values ${\cal A}$ by leveraging a comprehensive theorem database ${\cal T}$. 
Our framework supports three distinct reasoning paradigms:
(1) Symbolic reasoning employs search-based policies to identify and apply the most appropriate theorems from the knowledge base through systematic exploration.
(2) Neuro-symbolic reasoning integrates neural networks as theorem predictors while maintaining the logical rigor of formal deductive reasoning.
(3) Connectionist reasoning constructs end-to-end neural architectures that approximate the entire reasoning process as a black-box mapping function.
The first two approaches maintain strict adherence to formal logical deduction using theorems from the knowledge repository, while the connectionist approach relies on learned representations to approximate the reasoning process.

\subsection{Proposed Taxonomy}
A rigorous approach to automated GPS must address three core challenges:
\textbf{(1) Benchmark Construction:} \textit{What constitutes a well-defined geometric problem?}
\textbf{(2) Parser:} \textit{How can problems be formally interpreted?}
 \textbf{(3) Reasoner:} \textit{What systematic methods yield correct solutions?}
Our paradigm decomposes these into modular components (as illustrated in Fig.~\ref{fig:taxonomy}), with subsequent sections detailing their implementation:
Section~\ref{section:benchmark} introduces the Benchmark, which answers (i) by formalizing problem scope and diverse datasets. Section~\ref{section:parser} presents the parser, addressing (ii) via examining parsing methodologies including formal language space, diagram parser and text parser. Section~\ref{section:reason} discusses the reasoner, resolving (iii) through contrasting two dominant reasoning paradigms: ML frameworks and large language
model(LLM)-based systems.

\section{Benchmark Construction\label{section:benchmark}}
\begin{figure}[!t]
    \centering
    \includegraphics[width=\linewidth]{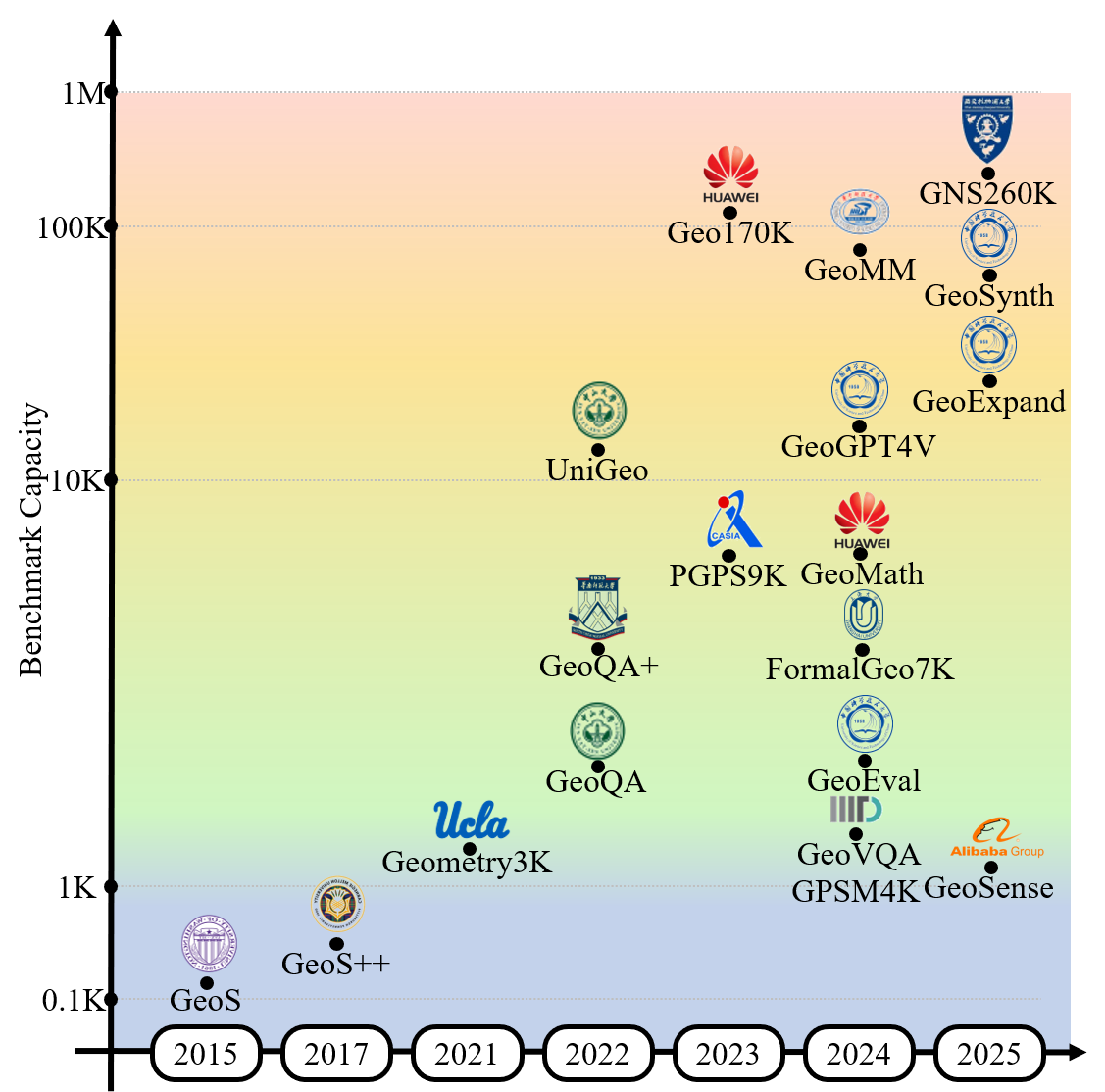}
    \vspace{-18pt}
    \caption{Research trend of benchmarks.}
    \label{fig:dataset}
\end{figure}

This section reviews the evolution of GPS benchmarks, as depicted in Fig.~\ref{fig:dataset}. 
Since 2015, benchmark development has focused on enhancing precision, expanding problem variety, and increasing overall scale. The advent of LLMs significantly accelerated this trend, primarily due to two factors: LLMs underscored the importance of data scale for superior performance, and they provided powerful tools for automated problem generation, transforming benchmark construction from a manual to a scalable process. Existing benchmarks are categorized by their construction technique: manual construction, LLM-assisted annotation, LLM-assisted augmentation, and LLM-assisted synthesis. Further details on these benchmarks are provided in Appendix \ref{section:appendix_benchmark}.

\subsection{Manual Construction\label{section:data_manu}}
GeoS built the first dataset for GPS \cite{seo2015solving}, covering 186 SAT plane geometry questions where every question has a textual description in English accompanied by a diagram and multiple choices. 
To further enhance the dataset scale, GeoS++ collected a total of 1,406 SAT style questions with more complex concepts \cite{sachan2017learning} across grades 6-10, while Geometry3K collected 3,002 problems from textbooks across grades 6-12 \cite{lu2021inter}.
Considering solution processes are also important in the GPS, PGPS9K added solution sequences as labels \cite{zhang2023multi} and constructed a dataset of 9,022.
To construct benchmarks with Chinese middle school exams, GeoQA \cite{chen2021geoqa} and advanced GeoQA+\cite{cao2022augmented} were proposed.
The annotated operations required to solve the problem in GeoQA are limited to a maximum of 4 steps, but that in GeoQA+ is with up to 8. 
To unify the proving problem with the calculation problem to construct the UniGeo benchmark \cite{chen2022unigeo}, UniGeo collects 9,543 proving problems and heritage 4,998 calculation problems from GeoQA.
Considering imperfect annotation limits the GPS to be SAT-level, Zhang \textit{et.al.} proposed a formal geometry theory and extended the annotation to the IMO-level \cite{zhang2024formal,zhang2023formalgeo}.
Thus, two datasets are constructed, i.e., FormalGeo7k and FormalGeo-IMO.
The former contains 6,981  geometry problems (SAT-level to IMO-level), while the latter includes 18 IMO-level challenging geometry problems. 
GeoEval integrates seven public datasets and newly collected geometry problems, covering areas such as plane geometry, solid geometry, and analytic geometry, to provide a benchmark for the GPS field \cite{zhang2024geoeval}. It consists of four subsets: GeoEval-2000, which includes 2,000 problems; GeoEval-backward, containing 750 problems designed for backward reasoning; GeoEval-aug, featuring 2,000 augmented problems with varied rephrasing; and GeoEval-hard, comprising 300 problems that focus on solid and analytic geometry.

\subsection{LLM-assisted Annotation\label{section:data_llm_annotation}}
LLMs exhibit dual capabilities in understanding multimodal geometric content and generating structured solution sequences, enabling their application in benchmark annotation.
Researchers leverage LLMs to generate step-by-step solution labels, considering that detailed reasoning descriptions significantly enhance human comprehension.
GPSM4K \cite{jaiswal2024advancing} and GeoVQA \cite{anand2024geovqa} constitute complementary benchmarks sourced from Indian mathematics textbooks (Grades 6–12), collectively containing approximately 4,400 calculation and proof problems. Both leverage Gemini Pro Vision to decompose textbook solutions into fundamental reasoning components, enhancing computational interpretability. 
GPSM4K employs Gemini Vision Pro for automated diagram captioning, enriching visual accessibility, while GeoVQA utilizes Gemini Vision Pro + Gemini Pro to dissect problems systematically, covering identifying key concepts, performing computations, and structuring solutions for multimodal question answering.
GeoSense \cite{xu2025geosense} employs LLMs to generate principle-level annotations for 1,789 geometric problems sourced from existing benchmarks and educational platforms. 
These annotations codify essential geometric knowledge into three classes, i.e., definitions, theorems, and formulas.
To ensure reliability, Xu \textit{et. al.} developed a semi-automated pipeline where GPT-4o and human experts collaboratively identify necessary principles, contextualize their application within geometric diagrams, and verify mathematical expressions for computational accuracy.
GNS260K addresses a critical gap in plane geometry datasets: the absence of natural language solution descriptions \cite{ning2025gns}. Leveraging GPT-4's advanced reasoning capabilities, it generates human-interpretable solving traces for 9,426 unique diagrams, scaled through augmentation to 18,852 knowledge prediction samples, 86,732 symbolic parsing samples, and 154,433 problem reasoning samples.
The annotation methodology employs a two-stage pipeline: (1) parsed problem clauses and symbolic solutions serve as structured input.
(2) GPT-4 converts formal solutions into pedagogical narratives using description-focused prompts.
This comprehensive annotation enables chain-of-thought reasoning in GPS systems, significantly enhancing both interpretability and educational applicability.

\subsection{LLM-assisted Augmentation \label{section:data_llm_aug}}
LLMs demonstrate significant generative capabilities for enhancing dataset quality. By automatically generating diverse variants and refining questions from existing data, LLMs can substantially improve a dataset’s diversity, complexity, and scale. This approach has become a widely adopted strategy for data augmentation.
GeoMath exemplifies this technique by utilizing GPT-3.5 for data augmentation through text rewriting and image caption generation. Specifically, it rephrases original problem statements in five distinct ways, expanding the sample size sixfold while increasing linguistic diversity. This method generates richer training examples, strengthening model generalization \cite{xu2024geo}.
Building on this paradigm, the Geo170K dataset employs ChatGPT to augment existing geometric datasets (GeoQA+ and Geometry3K).
Key augmentation strategies include (1) equation solving and value scaling, (2) re-formulating conditions as unknowns, and (3) systematic sentence paraphrasing.
Geo170K provides both alignment data (linking problems to geometric diagrams) and instruction data (problem-solving steps) derived from GeoQA+ and Geometry3K training sets \cite{xu2024geo}. The final dataset comprises approximately 60,000 geometric image-caption pairs and approximately 110,000 question-answer pairs, representing a significant scale advancement in geometric reasoning resources.

\subsection{LLM-assisted Synthesis \label{section:data_llm_syn}}
Researchers have increasingly leveraged LLMs for generating geometry problems due to their powerful capabilities. The primary goal is to utilize generative models to create accurate, non-contradictory problems, ensuring that the required theorems are clearly defined and the reasoning steps for problem-solving are controllable.
Image-generation oriented methods, exemplified by GeoGPT4V \cite{cai2024geogpt4v}, employ multi-modal fusion through a three-stage framework encompassing problem simplification, Wolfram code generation, and image scoring, utilizing 4.9K generated samples alongside 19K open-source data to achieve text-image correspondence for intuitive model training. 
Reverse reasoning methodologies, represented by GeomVerse \cite{kazemi2023geomverse} and GeoMM \cite{deng2024r}, adopt backward generation strategies that construct multi-hop reasoning problems through predefined geometric constructs, with GeoMM specifically implementing a Reverse Chain-of-Thought framework processing 20 geometric shapes across 87K samples to generate multi-step reasoning question-answer pairs. Symbolic reasoning approaches, notably AlphaGeometry \cite{trinh2024solving, chervonyi2025gold} and GeoGen \cite{pan2025enhancing}, leverage symbolic engines and traceback algorithms to derive mathematical facts from random geometric diagrams, with AlphaGeometry2 scaling to 0.3 billion training samples while GeoGen integrates symbolic reasoning with large language model capabilities, producing 45,526 question-answer pairs through GeoExpand and 129,230 initial diagrams via GeoSynth's predicate sampling methodology. Knowledge-directed generation, as demonstrated by GeoUni \cite{cheng2025geouni}, represents the methodological frontier by precisely generating geometric diagrams aligned with specific knowledge points, ensuring tight correspondence between textual problems and visual representations for customized educational applications. These methodological developments reveal a systematic evolution from elementary text-image pairing to sophisticated knowledge-directed generation, characterized by fundamental distinctions in generation strategies (forward versus reverse mechanisms), control mechanisms (difficulty adjustment versus knowledge targeting), data scalability (thousand-level samples to billion-scale datasets), and application domains (general-purpose generation versus educational customization), thereby establishing a comprehensive theoretical framework that advances geometric problem generation through diversified technical paradigms and scalable methodological innovations.

\section{Parser\label{section:parser}}
In our proposed paradigm, parsers $f$ are designed to map a geometry problem $({\cal X} \oplus {\cal Y}, {\cal Q})$ into a formal language representation $(\tilde{\cal X} \oplus \tilde{\cal Y}, \tilde{\cal Q})$, where the output belongs to a formal language space $\Omega$.
A formal language is a rigorously defined system of symbols and rules, commonly used in computer science and mathematics to describe and analyze structured information. 
In the context of GPS, formal languages enable the translation of geometric elements into symbolic expressions that support precise logical manipulation.
Compared to latent features, formal representations offer several advantages: they provide greater transparency and verifiability, facilitate robust symbolic reasoning, and allow for seamless integration of expert geometric knowledge \cite{seo2015solving, chen2021geoqa, trinh2024solving, chervonyi2025gold, zhang2023formalgeo}. 
In contrast, latent features often lack interpretability, struggle with rigorous logical inference, generalize poorly to novel structures, and face difficulties in incorporating domain expertise \cite{cao2022augmented, ning2025gns, zhang2023multi, xia2024geox}.
Hence, this paper focuses on surveying parsers that utilize formal languages, owing to their inherent advantages.

\subsection{Diagram Parser\label{section:paser_diagram}}
The development of diagram parsing techniques has shown significant advancement over time. 
GeoS pioneered this area using a publicly available diagram parser that provided confidence scores for literals and identified visual elements by maximizing pixel coverage, text-visual agreement, and element coherence \cite{seo2014diagram}. 
This approach could identify various shapes including lines, circles, and polygons, while serving dual purposes: computing diagram scores and extracting visual literals unavailable from text. 
Inter-GPS subsequently improved upon this foundation by developing a fully automatic parser without manual intervention \cite{lu2021inter}. This system applied Hough Transformation to extract geometric primitives, then employed RetinaNet for detecting diagram symbols and text regions, with MathPix handling optical character recognition. 
Unlike previous methods, Inter-GPS successfully managed special relational symbols such as parallel, perpendicular, and isosceles markings. 
PGDP further advanced the field through deep learning and graph reasoning \cite{hao2022pgdp5k}, introducing an end-to-end model (PGDPNet) that utilized modified instance segmentation for primitive extraction and graph neural networks for relation parsing and classification, incorporating both geometric features and prior knowledge. 
The most recent innovation, DFE-GPS-SigLIP, represents a multimodal approach integrating three key components \cite{zhang2025diagram}: a Diagram Formalizer, a Projection module, and a large language model. This system processes diagram features through SigLIP (Vision Encoder), aligns diverse inputs within the LLM's semantic space.

\subsection{Text Parser\label{section:paser_text}}

The development of text parsing techniques for geometry problems has undergone multiple evolutionary stages.
GeoS implemented a three-phase parsing pipeline: initially mapping textual terms to corresponding geometric concepts, subsequently identifying relationships between these concepts, and finally processing implications and coordinating conjunctions \cite{seo2014diagram}. GeoS employed a hypergraph structure to represent literals, where nodes corresponded to geometric concepts (constants, variables, functions, or predicates) and edges captured the relationships between concepts. The system first identified concepts within the text, then learned affinity scores for hypergraph edges, and ultimately completed relationships to satisfy type matching requirements in the formal language. 
GeoS++ enhanced this approach by adopting a part-based log-linear model that integrated multiple steps into a unified framework while maintaining a similar three-phase process \cite{sachan2017learning}. However, GeoS++ decomposed its model into representing concepts and representing relationships and then utilized a rule-based method similar to GeoS for relationship completion. 
Inter-GPS diverged from these approaches by employing template rules to transform problem text into formal language \cite{lu2021inter}. This system deliberately avoided sequence-to-sequence learning methods due to the limited scale of geometry datasets and the sensitivity of symbolic reasoning to noise, instead implementing rule-based parsing techniques with regular expressions to achieve more precise parsing results.

\section{Reasoner\label{section:reason}}

In this section, we introduce the reasoner, the central component in GPS systems. 
While Section~\ref{section:paradigm} details the three reasoning paradigms (symbolic, neuro-symbolic, and connectionist), this section classifies reasoners by their implementation tools: Machine Learning-based (ML-based) and LLM-based approaches.
Details are presented in Apppendix \ref{section:appendix_reasoner}

\subsection{ML-based Techniques\label{section:ml}}
\subsubsection{Rule-based Technique}
Rule-based techniques in GPS, often rooted in data mining, leverage the deductive nature of geometric reasoning. These approaches model geometry problems as logical systems: given values act as initial facts, geometric principles serve as rules, and the derivation of unknown values functions as logical deduction \cite{seo2015solving,sachan2017learning,zhang2024fgeo,zhang2024geoeval}.

GeoS pioneered this paradigm by translating geometry problems into logical expressions. Its solver utilized submodular optimization with greedy selection and basin-hopping combined with sequential least squares programming to maximize global constraint satisfaction \cite{seo2015solving}. Extending this foundation, GeoS++ parsed axiom information from mathematics textbooks and employed log-linear models to score axiom application sequences, effectively learning optimal theorem application strategies \cite{sachan2017learning}.
FGPS further advanced the field by incorporating symmetric problem-solving algorithms: forward search derives new conditions from initial ones until the goal is reached, while backward search decomposes goals into sub-goals until they match known conditions \cite{zhang2024fgeo,zhang2024formal}. This approach organizes the problem-solving process into a hypertree structure, where conditions are hypernodes and theorems are hyperedges. Experiments show that various search strategies (e.g., breadth-first, depth-first, random, and beam) offer differing performance benefits based on problem characteristics.

\subsubsection{NN-based Technique}
Recent research in computational geometry has integrated neural networks (NNs) into problem-solving frameworks through two distinct approaches: as theorem predictors within neuro-symbolic systems and as end-to-end reasoners in connectionist paradigms.

In the \textbf{neuro-symbolic paradigm}, NNs function as theorem predictors to guide symbolic reasoning. 
Inter-GPS exemplifies this approach by formulating GPS as goal-directed search, incorporating theorem knowledge as conditional rules for step-by-step symbolic reasoning \cite{lu2021inter}. The integration of a theorem predictor significantly enhances search efficiency by inferring likely theorem application sequences, addressing the limitations of brute-force enumeration strategies. 
Wu et al. advanced this paradigm with their Explainable Top-Down Problem Solver (TD-PS), which mimics human expert reasoning by starting from the target and working backward \cite{wu2024gps}. This approach employs target decomposition and condition processing mechanisms to ensure explainability while using neural-guided theorem prediction to constrain the search space effectively.

Alternatively, \textbf{ connectionist reasoning approaches} employ NNs as direct reasoners. 
GeoQA utilizes an LSTM decoder with attention mechanisms over multimodal information to generate sequential programs guided by embeddings \cite{chen2021geoqa}. 
PGPSNet further refines this approach by encoding diagram images with CNN and problem text with pre-trained language models \cite{hao2022pgdp5k}, fusing these modalities through bidirectional GRU encoders. 
A notable innovation in PGPSNet is its self-limited GRU decoder, which significantly reduces representation and search spaces to accelerate training and inference. This architecture employs simplified clauses describing basic relationships rather than complex multi-order logic forms, achieving performance comparable to more complex tree decoders but with substantially improved computational efficiency.

\subsection{LLM-based Techniques\label{section:llm}}


\subsubsection{In-context Learning}
In-Context Learning (ICL) has emerged as a significant capability of LLMs, enabling them to perform new tasks by conditioning on demonstration examples provided in the input prompt without explicit parameter updates. Recent research has leveraged this capability to address the complex domain of GPS through several innovative approaches.

G-LLaVA pioneered a method that concatenates mapped image features with text embeddings as input to LLMs, demonstrating superior performance over GPT4-V on geometry problems \cite{gao2023g}. 
Building on this foundation, DFE-GPS introduced a more structured approach by processing three input types, i.e., diagram features extracted by a Diagram Encoder, formal diagram language representations, and natural language inputs \cite{zhang2025diagram}. 
Further advancing the field, Geo-LLaVA enhanced performance through dataset augmentation and Retrieval Augmentation Generation (RAG) \cite{xu2024geo}, while GeoX employed a Generator-and-Sampler Transformer to create geometry content-aware queries and eliminate uninformative representations \cite{xia2024geox}. GeoX's formalized pre-training strategy demonstrated particular effectiveness in downstream geometry tasks through its generation of Minimal Sufficient Representations.
GeoUni introduced a significant methodological advancement with Geo-Reasoning-Adapter trained via a comprehensive reward function encompassing format, formalization, and accuracy components \cite{cheng2025geouni}. This approach substantially improved performance on multiple-choice and open-ended geometry questions. 
Departing from purely connectionist reasoning approaches, AlphaGeometry and AlphaGeometry2 positioned LLMs as auxiliary line predictors within a neuro-symbolic paradigm, using models like Gemini to predict auxiliary line construction before verification through symbolic reasoning \cite{trinh2024solving,chervonyi2025gold}.

\subsubsection{Chain of Thought}
Chain-of-Thought (CoT) prompting has emerged as a significant technique for enhancing language models' reasoning capabilities by generating intermediate steps before producing final answers. This approach is particularly valuable for geometric problem solving, which requires multi-hop mathematical reasoning across both textual and visual information. 

Research from GeomVerse demonstrates that while Vision-Language Models (VLMs) perform adequately on simpler problems, their effectiveness significantly diminishes with increased reasoning complexity \cite{kazemi2023geomverse}. However, training these models with CoT prompting substantially improves their geometric reasoning capabilities, especially when finetuned to generate both solutions and reasoning processes.

Several innovative approaches have extended the basic CoT paradigm. Wang et al. proposed a two-step zero-shot methodology that first generates a diagram graph to enhance the model's comprehension of geometric information, followed by problem-solving using this intermediate representation \cite{wang2024mutli}.
Ning et al. introduced GNS, a neural-symbolic framework that parses problems into symbolic clauses for explicit geometric comprehension and precise computation \cite{ning2025gns}. 
Further advancing this field, Yan et al. developed GeoQwen, conceptualizing the problem-solving process as a reasoning tree where each node represents a solution step \cite{shengyuan2024geo}. This approach implements backtracking when reaching impasses and employs a theorem predictor to identify relevant theorems, thereby reducing the search space and improving reasoning efficiency.

\subsubsection{Program of Thought}
Program-of-Thought (PoT) prompting is an advanced technique where large language models are guided to generate executable code, representing the reasoning steps required to solve a problem.
A persistent challenge with CoT finetuned Visual Language Models (VLMs) is their susceptibility to calculation errors and incorrect formula application, particularly in geometry problems requiring extended reasoning chains. 

To address these limitations, Sharma et al. introduced Geo-Coder, proposing modular code-finetuning as an alternative approach \cite{sharma2024geocoder}. This method generates Python code utilizing a predefined geometry library, offering three key advantages: deterministic calculations through code execution, reduced formula errors through predefined mathematical functions, and enhanced interpretability via templated print statements.
AlphaEvolve transforms geometric problems into algorithmic challenges aimed at finding the optimal geometric configurations under specific conditions \cite{Novikov2025AlphaEvolve}. It employs a multi-stage heuristic search approach to explore a vast solution space in search of the best algorithm to solve geometric problems.

Research by Duan et al. suggests that reasoning ability surpasses computational power in importance for geometric problem-solving. Recognizing the complementary strengths of CoT excelling in reasoning while PoT provides computational precision has led to hybrid methodologies \cite{duantowards}. 
The Reason-and-Execute (RaE) framework exemplifies this integration, embedding PoT within CoT as an execution mechanism. RaE employs reverse thinking to identify necessary geometric knowledge and logical steps while generating executable code blocks for precise arithmetic operations \cite{duan2024reason}.

\subsubsection{Chain of Agent}
Mouselinos et al. adopted simulacra-based conversational agents to construct Chain of Agent for GPS  \cite{mouselinos2024beyond}.In mathematical reasoning, a Chain of Agents could be employed by assigning one agent to parse and understand the word problem, another to identify the relevant mathematical operations or theorems, a third to perform the symbolic or numerical calculations, and a final agent to synthesize the steps and present the solution.
Studies suggest that the resulting cross-domain collaborative dialogue enables more sophisticated geometric problem-solving than traditional single-agent approaches by leveraging complementary cognitive strengths and distributing complex reasoning processes across specialized components.

\section{Challenges and Future Work\label{section:future}}
\textbf{Enhanced Multimodal Understanding and Logical Reasoning Capabilities}:
Future advancements will focus on two key LLM capabilities: (1) enhanced multimodal understanding of dynamic geometry \cite{goldenberg2012dynamic}, including motion analysis and property evolution tracking, and (2) robust logical reasoning for complex analytic geometry \cite{fischer2006complex}.

\textbf{Deep Fusion of Algebraic and Geometric Methods}:
Future LLM-powered GPS systems will integrate algebraic and geometric reasoning through three key capabilities: (1) automatic conversion of geometric problems into algebraic formulations, (2) application of symbolic computation tools (e.g., Wu's method \cite{wen1984decision}, Gröbner bases \cite{becker1993grobner}), and (3) geometric interpretation of algebraic results. 
This cross-paradigm approach is expected to facilitate breakthroughs in automated theorem proving, particularly for complex geometric inequalities that current methods struggle with.

\textbf{Creative Inductive Capabilities}:
Future LLMs will revolutionize geometric research by autonomously generating novel conjectures \cite{chen2025towards, li2025system} and discovering hidden patterns in complex structures.
By combining learned geometric knowledge with symbolic reasoning, these systems will serve as AI collaborators, proposing testable hypotheses, identifying new symmetries, and accelerating mathematical exploration beyond human intuition alone, while maintaining formal rigor.

\textbf{Self-Verification Ability}:
Causal reasoning modules can act as an internal monitor in the GPS task, flagging steps that violate fundamental spatial or relational expectations before errors propagate \cite{kiciman2023causal,wang2024causalbench}.
Integrating intuitive, causality-based plausibility checks can greatly enhance the robustness and reliability of automated GPS by enabling real-time detection and correction of reasoning errors, similar to human problem-solving.

\section{Conclusion}

This survey offers a comprehensive and systematic overview of GPS, tracing its developmental trajectory, especially within the era of large AI models. 
We establish a systematic paradigm and propose a structured taxonomy, classifying existing methodologies into three key technical perspectives: benchmark construction, parser, and reasoner, with each component further subdivided. Furthermore, we discuss current limitations and challenges within the field, identifying several promising directions for future research. 
This work provides valuable insights and a foundational framework to catalyze the advancement of GPS.

\bibliography{reference}

\begin{thebibliography}{60}
\providecommand{\natexlab}[1]{#1}
\providecommand{\url}[1]{\texttt{#1}}
\expandafter\ifx\csname urlstyle\endcsname\relax
  \providecommand{\doi}[1]{doi: #1}\else
  \providecommand{\doi}{doi: \begingroup \urlstyle{rm}\Url}\fi

\bibitem[Achiam et~al.(2023)Achiam, Adler, Agarwal, Ahmad, Akkaya, Aleman, Almeida, Altenschmidt, Altman, Anadkat, et~al.]{achiam2023gpt}
Achiam, J., Adler, S., Agarwal, S., Ahmad, L., Akkaya, I., Aleman, F.~L., Almeida, D., Altenschmidt, J., Altman, S., Anadkat, S., et~al.
\newblock Gpt-4 technical report.
\newblock \emph{arXiv preprint arXiv:2303.08774}, 2023.

\bibitem[Anand et~al.(2024)Anand, Jaiswal, Dharmadhikari, Marathe, Popat, Mital, Nair, Prasad, Kumar, Verma, et~al.]{anand2024geovqa}
Anand, A., Jaiswal, R., Dharmadhikari, A., Marathe, A., Popat, H., Mital, H., Nair, A.~R., Prasad, K., Kumar, S., Verma, A., et~al.
\newblock Geovqa: A comprehensive multimodal geometry dataset for secondary education.
\newblock In \emph{2024 IEEE 7th International Conference on Multimedia Information Processing and Retrieval (MIPR)}, pp.\  102--108. IEEE, 2024.

\bibitem[Awais et~al.(2024)Awais, Ahmed, Aslam, Rehman, Alamri, Bahaj, and Saba]{awais2024mathvision}
Awais, M., Ahmed, T., Aslam, M., Rehman, A., Alamri, F.~S., Bahaj, S.~A., and Saba, T.
\newblock Mathvision: An accessible intelligent agent for visually impaired people to understand mathematical equations.
\newblock \emph{IEEE Access}, 2024.

\bibitem[Bai et~al.(2023)Bai, Bai, Chu, Cui, Dang, Deng, Fan, Ge, Han, Huang, et~al.]{bai2023qwen}
Bai, J., Bai, S., Chu, Y., Cui, Z., Dang, K., Deng, X., Fan, Y., Ge, W., Han, Y., Huang, F., et~al.
\newblock Qwen technical report.
\newblock \emph{arXiv preprint arXiv:2309.16609}, 2023.

\bibitem[Becker et~al.(1993)Becker, Weispfenning, Becker, and Weispfenning]{becker1993grobner}
Becker, T., Weispfenning, V., Becker, T., and Weispfenning, V.
\newblock \emph{Gr{\"o}bner bases}.
\newblock Springer, 1993.

\bibitem[Bi \& Wang(2020)Bi and Wang]{bi2020computer}
Bi, Z. and Wang, X.
\newblock \emph{Computer aided design and manufacturing}.
\newblock John Wiley \& Sons, 2020.

\bibitem[Bobrow et~al.(1964)]{bobrow1964natural}
Bobrow, D. et~al.
\newblock Natural language input for a computer problem solving system.
\newblock 1964.

\bibitem[Cai et~al.(2024)Cai, Bao, Guo, Zhang, Song, and Zheng]{cai2024geogpt4v}
Cai, S., Bao, K., Guo, H., Zhang, J., Song, J., and Zheng, B.
\newblock Geogpt4v: Towards geometric multi-modal large language models with geometric image generation.
\newblock \emph{arXiv preprint arXiv:2406.11503}, 2024.

\bibitem[Cao \& Xiao(2022)Cao and Xiao]{cao2022augmented}
Cao, J. and Xiao, J.
\newblock An augmented benchmark dataset for geometric question answering through dual parallel text encoding.
\newblock In \emph{Proceedings of the 29th international conference on computational linguistics}, pp.\  1511--1520, 2022.

\bibitem[Chen et~al.(2021)Chen, Tang, Qin, Liang, Liu, Xing, and Lin]{chen2021geoqa}
Chen, J., Tang, J., Qin, J., Liang, X., Liu, L., Xing, E.~P., and Lin, L.
\newblock Geoqa: A geometric question answering benchmark towards multimodal numerical reasoning.
\newblock \emph{arXiv preprint arXiv:2105.14517}, 2021.

\bibitem[Chen et~al.(2022)Chen, Li, Qin, Lu, Lin, Chen, and Liang]{chen2022unigeo}
Chen, J., Li, T., Qin, J., Lu, P., Lin, L., Chen, C., and Liang, X.
\newblock Unigeo: Unifying geometry logical reasoning via reformulating mathematical expression.
\newblock \emph{arXiv preprint arXiv:2212.02746}, 2022.

\bibitem[Chen et~al.(2025)Chen, Qin, Liu, Peng, Guan, Wang, Hu, Zhou, Gao, and Che]{chen2025towards}
Chen, Q., Qin, L., Liu, J., Peng, D., Guan, J., Wang, P., Hu, M., Zhou, Y., Gao, T., and Che, W.
\newblock Towards reasoning era: A survey of long chain-of-thought for reasoning large language models.
\newblock \emph{arXiv preprint arXiv:2503.09567}, 2025.

\bibitem[Cheng et~al.(2025)Cheng, Zhang, Chen, Deng, Qin, and Ma]{cheng2025geouni}
Cheng, J.-K., Zhang, Z., Chen, R., Deng, J., Qin, Z., and Ma, J.
\newblock Geouni: A unified model for generating geometry diagrams, problems and problem solutions.
\newblock \emph{arXiv preprint arXiv:2504.10146}, 2025.

\bibitem[Chervonyi et~al.(2025)Chervonyi, Trinh, Ol{\v{s}}{\'a}k, Yang, Nguyen, Menegali, Jung, Verma, Le, and Luong]{chervonyi2025gold}
Chervonyi, Y., Trinh, T.~H., Ol{\v{s}}{\'a}k, M., Yang, X., Nguyen, H., Menegali, M., Jung, J., Verma, V., Le, Q.~V., and Luong, T.
\newblock Gold-medalist performance in solving olympiad geometry with alphageometry2.
\newblock \emph{arXiv preprint arXiv:2502.03544}, 2025.

\bibitem[Chou et~al.(1996)Chou, Gao, and Zhang]{chou1996automated}
Chou, S.-C., Gao, X.-S., and Zhang, J.-Z.
\newblock Automated generation of readable proofs with geometric invariants: Ii. theorem proving with full-angles.
\newblock \emph{Journal of Automated Reasoning}, 17\penalty0 (3):\penalty0 349--370, 1996.

\bibitem[Deng et~al.(2024)Deng, Liu, Li, Luo, Wu, Zhang, Lyu, Zhang, Zhang, Ding, et~al.]{deng2024r}
Deng, L., Liu, Y., Li, B., Luo, D., Wu, L., Zhang, C., Lyu, P., Zhang, Z., Zhang, G., Ding, E., et~al.
\newblock R-cot: Reverse chain-of-thought problem generation for geometric reasoning in large multimodal models.
\newblock \emph{arXiv preprint arXiv:2410.17885}, 2024.

\bibitem[Duan et~al.()Duan, Tan, Fang, Guan, Zhou, Huang, Gong, and Huang]{duantowards}
Duan, X., Tan, D., Fang, L., Guan, Q., Zhou, Y., Huang, X., Gong, Z., and Huang, C.
\newblock Towards geometry problems solving employing gpt-4 vision with few-shot prompting: An empirical study of what matters.

\bibitem[Duan et~al.(2024)Duan, Tan, Fang, Zhou, He, Chen, Wu, Chen, Gong, Luo, et~al.]{duan2024reason}
Duan, X., Tan, D., Fang, L., Zhou, Y., He, C., Chen, Z., Wu, L., Chen, G., Gong, Z., Luo, W., et~al.
\newblock Reason-and-execute prompting: Enhancing multi-modal large language models for solving geometry questions.
\newblock In \emph{Proceedings of the 32nd ACM International Conference on Multimedia}, pp.\  6959--6968, 2024.

\bibitem[Fischer(2006)]{fischer2006complex}
Fischer, G.
\newblock \emph{Complex analytic geometry}, volume 538.
\newblock Springer, 2006.

\bibitem[Gao et~al.(2023)Gao, Pi, Zhang, Ye, Zhong, Wang, Hong, Han, Xu, Li, et~al.]{gao2023g}
Gao, J., Pi, R., Zhang, J., Ye, J., Zhong, W., Wang, Y., Hong, L., Han, J., Xu, H., Li, Z., et~al.
\newblock G-llava: Solving geometric problem with multi-modal large language model.
\newblock \emph{arXiv preprint arXiv:2312.11370}, 2023.

\bibitem[Goldenberg \& Cuoco(2012)Goldenberg and Cuoco]{goldenberg2012dynamic}
Goldenberg, E.~P. and Cuoco, A.~A.
\newblock What is dynamic geometry?
\newblock In \emph{Designing learning environments for developing understanding of geometry and space}, pp.\  351--367. Routledge, 2012.

\bibitem[Grattafiori et~al.(2024)Grattafiori, Dubey, Jauhri, Pandey, Kadian, Al-Dahle, Letman, Mathur, Schelten, Vaughan, et~al.]{grattafiori2024llama}
Grattafiori, A., Dubey, A., Jauhri, A., Pandey, A., Kadian, A., Al-Dahle, A., Letman, A., Mathur, A., Schelten, A., Vaughan, A., et~al.
\newblock The llama 3 herd of models.
\newblock \emph{arXiv preprint arXiv:2407.21783}, 2024.

\bibitem[Hao et~al.(2022)Hao, Zhang, Yin, and Huang]{hao2022pgdp5k}
Hao, Y., Zhang, M., Yin, F., and Huang, L.-L.
\newblock Pgdp5k: A diagram parsing dataset for plane geometry problems.
\newblock In \emph{2022 26th international conference on pattern recognition (ICPR)}, pp.\  1763--1769. IEEE, 2022.

\bibitem[Itzhaky et~al.(2013)Itzhaky, Gulwani, Immerman, and Sagiv]{itzhaky2013solving}
Itzhaky, S., Gulwani, S., Immerman, N., and Sagiv, M.
\newblock Solving geometry problems using a combination of symbolic and numerical reasoning.
\newblock In \emph{Logic for Programming, Artificial Intelligence, and Reasoning: 19th International Conference, LPAR-19, Stellenbosch, South Africa, December 14-19, 2013. Proceedings 19}, pp.\  457--472. Springer, 2013.

\bibitem[Jaiswal et~al.(2024)Jaiswal, Anand, and Shah]{jaiswal2024advancing}
Jaiswal, R., Anand, A., and Shah, R.~R.
\newblock Advancing multimodal llms: A focus on geometry problem solving reasoning and sequential scoring.
\newblock In \emph{Proceedings of the 6th ACM International Conference on Multimedia in Asia}, pp.\  1--7, 2024.

\bibitem[Kazemi et~al.(2023)Kazemi, Alvari, Anand, Wu, Chen, and Soricut]{kazemi2023geomverse}
Kazemi, M., Alvari, H., Anand, A., Wu, J., Chen, X., and Soricut, R.
\newblock Geomverse: A systematic evaluation of large models for geometric reasoning.
\newblock \emph{arXiv preprint arXiv:2312.12241}, 2023.

\bibitem[Kiciman et~al.(2023)Kiciman, Ness, Sharma, and Tan]{kiciman2023causal}
Kiciman, E., Ness, R., Sharma, A., and Tan, C.
\newblock Causal reasoning and large language models: Opening a new frontier for causality.
\newblock \emph{Transactions on Machine Learning Research}, 2023.

\bibitem[Li et~al.(2025)Li, Zhang, Zhang, Zhang, Liu, Yao, Xu, Zheng, Wang, Chen, et~al.]{li2025system}
Li, Z.-Z., Zhang, D., Zhang, M.-L., Zhang, J., Liu, Z., Yao, Y., Xu, H., Zheng, J., Wang, P.-J., Chen, X., et~al.
\newblock From system 1 to system 2: A survey of reasoning large language models.
\newblock \emph{arXiv preprint arXiv:2502.17419}, 2025.

\bibitem[Lin et~al.(2023)Lin, Huang, and Lu]{lin2023artificial}
Lin, C.-C., Huang, A.~Y., and Lu, O.~H.
\newblock Artificial intelligence in intelligent tutoring systems toward sustainable education: a systematic review.
\newblock \emph{Smart Learning Environments}, 10\penalty0 (1):\penalty0 41, 2023.

\bibitem[Liu et~al.(2023)Liu, Li, Wu, and Lee]{liu2023visual}
Liu, H., Li, C., Wu, Q., and Lee, Y.~J.
\newblock Visual instruction tuning.
\newblock \emph{Advances in neural information processing systems}, 36:\penalty0 34892--34916, 2023.

\bibitem[Liu et~al.(2025)Liu, Qiu, Li, Dai, Zhu, Hu, Yang, and King]{liu2025survey}
Liu, J., Qiu, Z., Li, Z., Dai, Q., Zhu, J., Hu, M., Yang, M., and King, I.
\newblock A survey of personalized large language models: Progress and future directions.
\newblock \emph{arXiv preprint arXiv:2502.11528}, 2025.

\bibitem[Loveland(2016)]{loveland2016automated}
Loveland, D.~W.
\newblock \emph{Automated theorem proving: A logical basis}.
\newblock Elsevier, 2016.

\bibitem[Lu et~al.(2021)Lu, Gong, Jiang, Qiu, Huang, Liang, and Zhu]{lu2021inter}
Lu, P., Gong, R., Jiang, S., Qiu, L., Huang, S., Liang, X., and Zhu, S.-C.
\newblock Inter-gps: Interpretable geometry problem solving with formal language and symbolic reasoning.
\newblock \emph{arXiv preprint arXiv:2105.04165}, 2021.

\bibitem[Lu et~al.(2023)Lu, Bansal, Xia, Liu, Li, Hajishirzi, Cheng, Chang, Galley, and Gao]{lu2023mathvista}
Lu, P., Bansal, H., Xia, T., Liu, J., Li, C., Hajishirzi, H., Cheng, H., Chang, K.-W., Galley, M., and Gao, J.
\newblock Mathvista: Evaluating mathematical reasoning of foundation models in visual contexts.
\newblock \emph{arXiv preprint arXiv:2310.02255}, 2023.

\bibitem[Mouselinos et~al.(2024)Mouselinos, Michalewski, and Malinowski]{mouselinos2024beyond}
Mouselinos, S., Michalewski, H., and Malinowski, M.
\newblock Beyond lines and circles: Unveiling the geometric reasoning gap in large language models.
\newblock \emph{arXiv preprint arXiv:2402.03877}, 2024.

\bibitem[Ning et~al.(2025)Ning, Zhou, Wang, Huang, and Huang]{ning2025gns}
Ning, M., Zhou, Z., Wang, Q., Huang, X., and Huang, K.
\newblock Gns: Solving plane geometry problems by neural-symbolic reasoning with multi-modal llms.
\newblock In \emph{Proceedings of the AAAI Conference on Artificial Intelligence}, volume~39, pp.\  24957--24965, 2025.

\bibitem[Novikov et~al.(2025)Novikov, Vu, Eisenberger, Dupont, Huang, Wagner, Shirobokov, Kozlovskii, Ruiz, Mehrabian, Kumar, See, Chaudhuri, Holland, Davies, Nowozin, Kohli, and Balog]{Novikov2025AlphaEvolve}
Novikov, A., Vu, N., Eisenberger, M., Dupont, E., Huang, P.-S., Wagner, A.~Z., Shirobokov, S., Kozlovskii, B., Ruiz, F. J.~R., Mehrabian, A., Kumar, M.~P., See, A., Chaudhuri, S., Holland, G., Davies, A., Nowozin, S., Kohli, P., and Balog, M.
\newblock Alphaevolve: A coding agent for scientific and algorithmic discovery.
\newblock \emph{Google DeepMind}, 2025.
\newblock URL \url{https://colab.research.google.com/github/google-deepmind/alphaevolve_results/blob/master/mathematical_results.ipynb}.
\newblock Accessed: 2025-05-25.

\bibitem[Pan et~al.(2025)Pan, Zhang, Hu, Ma, Du, Zhang, Liu, Gao, and Ma]{pan2025enhancing}
Pan, Y., Zhang, Z., Hu, P., Ma, J., Du, J., Zhang, J., Liu, Q., Gao, J., and Ma, F.
\newblock Enhancing the geometric problem-solving ability of multimodal llms via symbolic-neural integration.
\newblock \emph{arXiv preprint arXiv:2504.12773}, 2025.

\bibitem[Sachan \& Xing(2017)Sachan and Xing]{sachan2017learning}
Sachan, M. and Xing, E.
\newblock Learning to solve geometry problems from natural language demonstrations in textbooks.
\newblock In \emph{Proceedings of the 6th joint conference on lexical and computational semantics (* SEM 2017)}, pp.\  251--261, 2017.

\bibitem[Seo et~al.(2015)Seo, Hajishirzi, Farhadi, Etzioni, and Malcolm]{seo2015solving}
Seo, M., Hajishirzi, H., Farhadi, A., Etzioni, O., and Malcolm, C.
\newblock Solving geometry problems: Combining text and diagram interpretation.
\newblock In \emph{Proceedings of the 2015 conference on empirical methods in natural language processing}, pp.\  1466--1476, 2015.

\bibitem[Seo et~al.(2014)Seo, Hajishirzi, Farhadi, and Etzioni]{seo2014diagram}
Seo, M.~J., Hajishirzi, H., Farhadi, A., and Etzioni, O.
\newblock Diagram understanding in geometry questions.
\newblock In \emph{Proceedings of the AAAI Conference on Artificial Intelligence}, volume~28, 2014.

\bibitem[Sharma et~al.(2024)Sharma, Dalmia, Kazemi, Zouaq, and Pal]{sharma2024geocoder}
Sharma, A., Dalmia, A., Kazemi, M., Zouaq, A., and Pal, C.~J.
\newblock Geocoder: Solving geometry problems by generating modular code through vision-language models.
\newblock \emph{arXiv preprint arXiv:2410.13510}, 2024.

\bibitem[Tang et~al.(2024)Tang, Zhang, Zhu, and Liu]{tang2024tangram}
Tang, J., Zhang, C., Zhu, X., and Liu, M.
\newblock Tangram: A challenging benchmark for geometric element recognizing.
\newblock \emph{arXiv preprint arXiv:2408.13854}, 2024.

\bibitem[Trinh et~al.(2024)Trinh, Wu, Le, He, and Luong]{trinh2024solving}
Trinh, T.~H., Wu, Y., Le, Q.~V., He, H., and Luong, T.
\newblock Solving olympiad geometry without human demonstrations.
\newblock \emph{Nature}, 625\penalty0 (7995):\penalty0 476--482, 2024.

\bibitem[Wang \& Ma(2024)Wang and Ma]{wang2024mutli}
Wang, L. and Ma, J.
\newblock Mutli-step chain-of-thought in geometry problem solving.
\newblock In \emph{2024 4th International Conference on Electronic Information Engineering and Computer Science (EIECS)}, pp.\  1113--1117. IEEE, 2024.

\bibitem[Wang(2024)]{wang2024causalbench}
Wang, Z.
\newblock Causalbench: A comprehensive benchmark for evaluating causal reasoning capabilities of large language models.
\newblock In \emph{Proceedings of the 10th SIGHAN Workshop on Chinese Language Processing (SIGHAN-10)}, pp.\  143--151, 2024.

\bibitem[Wen-Ts{\"u}n(1984)]{wen1984decision}
Wen-Ts{\"u}n, W.
\newblock On the decision problem and the mechanization of theorem-proving in elementary geometry$^1$.
\newblock \emph{Automated Theorem Proving: After 25 Years: After 25 Years}, 89:\penalty0 213, 1984.

\bibitem[Wu et~al.(2024)Wu, Zhang, Liu, Tang, Wang, Wang, and Wang]{wu2024gps}
Wu, W., Zhang, L., Liu, J., Tang, X., Wang, Y., Wang, S., and Wang, Q.
\newblock E-gps: Explainable geometry problem solving via top-down solver and bottom-up generator.
\newblock In \emph{Proceedings of the IEEE/CVF Conference on Computer Vision and Pattern Recognition}, pp.\  13828--13837, 2024.

\bibitem[Xia et~al.(2024)Xia, Li, Ye, Wu, Zhou, Yuan, Peng, Cai, Yan, Wang, et~al.]{xia2024geox}
Xia, R., Li, M., Ye, H., Wu, W., Zhou, H., Yuan, J., Peng, T., Cai, X., Yan, X., Wang, B., et~al.
\newblock Geox: Geometric problem solving through unified formalized vision-language pre-training.
\newblock \emph{arXiv preprint arXiv:2412.11863}, 2024.

\bibitem[Xu et~al.(2025)Xu, Zhao, Wang, Wang, Pi, Wang, Zhang, Gu, Li, Zhu, et~al.]{xu2025geosense}
Xu, L., Zhao, Y., Wang, J., Wang, Y., Pi, B., Wang, C., Zhang, M., Gu, J., Li, X., Zhu, X., et~al.
\newblock Geosense: Evaluating identification and application of geometric principles in multimodal reasoning.
\newblock \emph{arXiv preprint arXiv:2504.12597}, 2025.

\bibitem[Xu et~al.(2024)Xu, Luo, and Shi]{xu2024geo}
Xu, S., Luo, Y., and Shi, W.
\newblock Geo-llava: A large multi-modal model for solving geometry math problems with meta in-context learning.
\newblock In \emph{Proceedings of the 2nd Workshop on Large Generative Models Meet Multimodal Applications}, pp.\  11--15, 2024.

\bibitem[Yan \& Zhong(2024)Yan and Zhong]{shengyuan2024geo}
Yan, S. and Zhong, X.
\newblock Geo-qwen: A geometry problem-solving method based on generative large language models and heuristic reasoning.
\newblock In \emph{2024 21st International Computer Conference on Wavelet Active Media Technology and Information Processing (ICCWAMTIP)}, pp.\  1--9. IEEE, 2024.

\bibitem[Zhang et~al.(2024{\natexlab{a}})Zhang, Li, Zhang, Yin, Liu, and Moshfeghi]{zhang2024geoeval}
Zhang, J., Li, Z., Zhang, M., Yin, F., Liu, C., and Moshfeghi, Y.
\newblock Geoeval: benchmark for evaluating llms and multi-modal models on geometry problem-solving.
\newblock \emph{arXiv preprint arXiv:2402.10104}, 2024{\natexlab{a}}.

\bibitem[Zhang et~al.(2022)Zhang, Yin, Hao, and Liu]{zhang2022plane}
Zhang, M.-L., Yin, F., Hao, Y.-H., and Liu, C.-L.
\newblock Plane geometry diagram parsing.
\newblock \emph{arXiv preprint arXiv:2205.09363}, 2022.

\bibitem[Zhang et~al.(2023{\natexlab{a}})Zhang, Yin, and Liu]{zhang2023multi}
Zhang, M.-L., Yin, F., and Liu, C.-L.
\newblock A multi-modal neural geometric solver with textual clauses parsed from diagram.
\newblock \emph{arXiv preprint arXiv:2302.11097}, 2023{\natexlab{a}}.

\bibitem[Zhang et~al.(2024{\natexlab{b}})Zhang, Jiang, Zhang, Lin, Guo, Qiu, Zhou, Lu, Chang, Qiao, et~al.]{zhang2024mathverse}
Zhang, R., Jiang, D., Zhang, Y., Lin, H., Guo, Z., Qiu, P., Zhou, A., Lu, P., Chang, K.-W., Qiao, Y., et~al.
\newblock Mathverse: Does your multi-modal llm truly see the diagrams in visual math problems?
\newblock In \emph{European Conference on Computer Vision}, pp.\  169--186. Springer, 2024{\natexlab{b}}.

\bibitem[Zhang et~al.(2023{\natexlab{b}})Zhang, Zhu, He, Zou, Huang, Jin, Guo, Mao, Li, Zhu, et~al.]{zhang2023formalgeo}
Zhang, X., Zhu, N., He, Y., Zou, J., Huang, Q., Jin, X., Guo, Y., Mao, C., Li, Y., Zhu, Z., et~al.
\newblock Formalgeo: An extensible formalized framework for olympiad geometric problem solving.
\newblock \emph{arXiv preprint arXiv:2310.18021}, 2023{\natexlab{b}}.

\bibitem[Zhang et~al.(2024{\natexlab{c}})Zhang, Zhu, He, Zou, Qin, Li, and Leng]{zhang2024fgeo}
Zhang, X., Zhu, N., He, Y., Zou, J., Qin, C., Li, Y., and Leng, T.
\newblock Fgeo-sss: A search-based symbolic solver for human-like automated geometric reasoning.
\newblock \emph{Symmetry}, 16\penalty0 (4):\penalty0 404, 2024{\natexlab{c}}.

\bibitem[Zhang et~al.(2024{\natexlab{d}})Zhang, Zhu, Qin, Yang, Zeng, and Leng]{zhang2024formal}
Zhang, X., Zhu, N., Qin, C., Yang, L., Zeng, Z., and Leng, T.
\newblock Formal representation and solution of plane geometric problems.
\newblock In \emph{The 4th Workshop on Mathematical Reasoning and AI at NeurIPS'24}, 2024{\natexlab{d}}.

\bibitem[Zhang et~al.(2025)Zhang, Cheng, Deng, Tian, Ma, Qin, Zhang, Zhu, and Leng]{zhang2025diagram}
Zhang, Z., Cheng, J.-K., Deng, J., Tian, L., Ma, J., Qin, Z., Zhang, X., Zhu, N., and Leng, T.
\newblock Diagram formalization enhanced multi-modal geometry problem solver.
\newblock In \emph{ICASSP 2025-2025 IEEE International Conference on Acoustics, Speech and Signal Processing (ICASSP)}, pp.\  1--5. IEEE, 2025.

\end{thebibliography}
\bibliographystyle{icml2025}

\section{Appendix}
\subsection{Details for Benchmarks\label{section:appendix_benchmark}}
\begin{table*}[h]
    \centering
    \caption{Benchmarks in geometry problem solving.}
    \label{tab:benchmark}
    \rowcolors{1}{white}{hidden-blue!30}
    \begin{tabular}{m{2cm}<{\centering} m{1.5cm}<{\centering} m{2cm}<{\centering} m{2cm}<{\centering} m{3cm}<{\centering} m{4cm}<{\centering}}
        \toprule
        Benchmark & Scale &Level & Type & Format  & Source \\
        \midrule
        GeoS     & 186 & \thead{SAT-level \\ (Grades 6-10)} & Calculation &Choice-Answer & SAT\\
        GeoS++     & 1,406 & \thead{SAT-level \\ (Grades 6-10)} & Calculation &Choice-Answer & SAT \& Textbook\\
        Geometry3K & 3,002 & \thead{SAT-level \\ (Grades 6-12)} & Calculation &Choice-Answer & Textbook\\
        PGPS9K & 9,022 &\thead{SAT-level \\ (Grades 6-12)} & Calculation & Solution Sequence & Textbook\\          
        GeoQA & 4,998 &\thead{SAT-level \\ (Grades 6-12)} & Calculation & Key point \& Solution Sequence &  Exams (Chinese) \\
        GeoQA+ & 7,528 &\thead{SAT-level \\ (Grades 6-12)} & Calculation & Key point \& Solution Sequence & Exams (Chinese) \\
        UniGeo & 1,7071 &\thead{SAT-level \\ (Grades 6-12)} & Calculation \& Proving & Solution Sequence  & Online Education Website \\
        FormalGeo7K & 6,987 & SAT-level to IMO-level & Calculation \& Proving & Solution Sequence  & Online Website\\   
        GeoEval-2000  & 2,000 & SAT-level & Calculation \& Proving & Solution Sequence  & \thead{Geometry3K, PGPS9K, \\ UniGeo, MATH, GeoQA+, \\ GeometryQA, MathQA}\\ 
        GeoEval-backward  & 750 & SAT-level & Calculation \& Proving & \thead{Solution Sequence \\ (Backward Reasoning)} & \thead{Geometry3K, PGPS9K, \\ UniGeo, MATH, GeoQA+, \\ GeometryQA, MathQA}\\ 
        GeoEval-aug  & 2,000 & SAT-level & Calculation \& Proving & Solution Sequence  & \thead{Geometry3K, PGPS9K, \\ UniGeo, MATH, GeoQA+, \\ GeometryQA, MathQA}\\ 
        GeoEval-hard  & 300 & SAT-level & Calculation \& Proving & Solution Sequence  & \thead{Geometry3K, PGPS9K, \\ UniGeo, MATH, GeoQA+, \\ GeometryQA, MathQA}\\ 
        GPSM4K \& GeoVQA & 4,440 &\thead{SAT-level \\ (Grades 6-12)} & Calculation \& Proving & Solution Sequence  & Textbook (Indian) \\
        GeoSense & 1,789 & SAT-level & Calculation & Solution Sequence & \thead{Online Website \\ (English \& Chinese))} \\
        GNS260K & 154,433 &  \thead{SAT-level \\ (Grades 6-12)} & Calculation & Solution Sequence & \thead{Synthesis \\ (PGPS9K,GeoQA+,Geo170K)}\\
        GeoMath & 9,155 & SAT-level & Calculation \& Proving & Solution Sequence  & Online Education Website (Chinese)\\
        Geo170K & 110,000 & \thead{SAT-level \\ (Grades 6-12)} & Calculation & Solution Sequence & \thead{Synthesis \\ (Geometry3K, GeoQA+)} \\
        GeoGPT4V & 23,955 & \thead{SAT-level \\ (Grades 6-12)} & Calculation & Solution Sequence & Synthesis (UniGeo-Calculation, Geometry3K, GeoQA+)\\
        GeoMM & 87,000 & SAT-level & Calculation & Solution Sequence & Synthesis\\
        GeoExpand & 45,526 & SAT-level & Calculation & Solution Sequence & \thead{ Synthesis \\ (Geometry3K and PGPS9K)} \\
        GeoSynth & 62,868 & SAT-level & Calculation & Solution Sequence & Synthesis\\
        \bottomrule
    \end{tabular}
\end{table*}
Detailed information about benchmarks is presented in Table.~\ref{tab:benchmark}.
\subsection{Details for Reasoners\label{section:appendix_reasoner}}
\begin{table*}[h]
    \centering
    \caption{Reasoners in geometry problem solving.}
    \label{tab:reasoner}
    \begin{tabular}{p{3.2cm}<{\centering} p{1.5cm}<{\centering} p{2cm}<{\centering}| p{3.2cm}<{\centering} p{1.5cm}<{\centering} p{2cm}<{\centering}}
        \toprule
        Reasoner &Technique & Paradigm & Reasoner &Technique & Paradigm \\
        \midrule
        \rowcolor{hidden-blue!30} \multicolumn{6}{c}{\textbf{ML-based Techniques}} \\
        \thead{GeoS \\ \cite{seo2015solving}} & \thead{Rule-based}& Symbolic & 
        \thead{GeoS+ \\ \cite{sachan2017learning}} & \thead{Rule-based}& Symbolic\\
        \thead{FGPS \\ \cite{zhang2024fgeo}} & \thead{Rule-based}& Symbolic&        
        \thead{InterGPS \\ \cite{lu2021inter}} & \thead{NN-based} & Neuro-Symbolic \\
        \thead{E-GPS \\ \cite{wu2024gps}} & \thead{NN-based} & Neuro-Symbolic & 
        \thead{GeoQA \\ \cite{chen2021geoqa}} & \thead{NN-based} & Connectionist \\
        \thead{PGPSNet \\ \cite{hao2022pgdp5k}} & \thead{NN-based} & Connectionist &  &  &  \\

        \rowcolor{hidden-blue!30} \multicolumn{6}{c}{\textbf{LLM-based Techniques}} \\
        \thead{G-LLaVA \\ \cite{gao2023g}} & \thead{ICL} & Connectionist & 
        \thead{DFE-GPS \\ \cite{zhang2025diagram}} & \thead{ICL} & Connectionist \\
        \thead{Geo-LLaVA \\ \cite{xu2024geo}} & \thead{ICL} & Connectionist & 
        \thead{GeoX \\ \cite{xia2024geox}} & \thead{ICL} & Connectionist \\
        \thead{GeoUni \\ \cite{cheng2025geouni}} & \thead{ICL} & Connectionist & 
        \thead{AlphaGeometry \\ \cite{trinh2024solving}} & \thead{ICL} & Neuro-Symbolic \\
        \thead{AlphaGeometry2 \\ \cite{chervonyi2025gold}} & \thead{ICL} & Neuro-Symbolic & 
        \thead{GeomVerse \\ \cite{kazemi2023geomverse}} & \thead{CoT} & Connectionist \\
        \thead{Two-Stage CoT \\ \cite{wang2024mutli}} & \thead{CoT} & Connectionist & 
        \thead{GNS \\ \cite{ning2025gns}} & \thead{CoT} & Connectionist \\       
        \thead{GeoQwen \\ \cite{shengyuan2024geo}} & \thead{CoT} & Connectionist &   
        \thead{GeoCoder \\ \cite{sharma2024geocoder}} & \thead{PoT} &  Connectionist\\ 
        \thead{RaE \\ \cite{duan2024reason}} & \thead{PoT} &  Connectionist& 
        \thead{Simulacra-\\Conversational Agents \\ \cite{mouselinos2024beyond}} & \thead{CoA} &  Connectionist\\ 
        \bottomrule
    \end{tabular}
\end{table*}
Detailed information about benchmarks is presented in Table.~\ref{tab:reasoner}.

\subsection{Related Benchmarks}
\textbf{Benchmarks for Diagram Parsing:}
Since diagrams play an important role in GPS, many researchers concentrate on constructing benchmarks for geometry diagrams with fine-grained annotation.
The PGDP5K dataset contains 5,000 diagram samples \cite{hao2022pgdp5k}, consisting of 1,813 non-duplicated images from the Geometry3K dataset and other 3,187 images collected from three popular textbooks across grades 6-12 on mathematics website. 
In contrast to previous datasets, diagrams in PGDP5K have more complex layouts such as multiple classes of primitives and complicated primitive relations, which make our dataset more challenging.
Tangram, includes 1,080 diverse geometric diagrams sourced from primary and secondary school exams, competitions, and textbooks, covering from simple basic geometric shapes to complex combinations, aiming to evaluate the performance of LMMs on geometric element recognition \cite{tang2024tangram}. 
Each diagram is associated with four questions about the diagram.
SynthGeo228K is a collect of over 228,000 geometric diagrams, generated from a single template by modifying point positions and orientations, with rotation being a common data augmentation method \cite{zhang2025diagram}.
SynthGeo228K, comprising 462 templates, provides a comprehensive exploration of basic geometry through the relationships between points, lines, and triangles, polygons, and circles. 

\textbf{Benchmarks for Mathematical Visual Reasoning:}
Recent advances in multimodal learning have produced several specialized datasets for mathematical visual reasoning. MathVista \cite{lu2023mathvista}, MathVerse \cite{zhang2024mathverse}, and MathVision \cite{awais2024mathvision} represent significant contributions to this field, all designed to evaluate multimodal models' performance through rich metadata annotations. These datasets differ primarily in their data collection approaches: MathVista integrates existing resources, MathVerse adapts publicly available materials, and MathVision collects entirely new samples from mathematical competitions. MathVista comprises 6,141 samples with detailed metadata annotations including question types, answer types, and task categories. MathVerse contains 2,612 high-quality mathematical problems with diagrams, each transformed into six variants with varying information distribution across modalities: text-dominant, text-lite, text-only, vision-intensive, vision-dominant, and vision-only. MathVision features 3,040 mathematics-visual question-answer pairs, meticulously curated through a four-stage filtering process by ten university students. 
Besides, math reasoning with visual content is also an interesting topic.

\end{document}